\documentclass[lettersize,journal]{IEEEtran}
\usepackage{amsmath,amsfonts}
\usepackage{algorithmic}
\usepackage{algorithm}
\usepackage{array}
\usepackage{textcomp}
\usepackage{stfloats}
\usepackage{url}
\usepackage{verbatim}
\usepackage{graphicx}
\usepackage{cite}
\usepackage{orcidlink}
\usepackage{multirow} 
\usepackage{booktabs}
\usepackage{balance}

\begin{document}

\title{DepthVision: Enabling Robust Vision–Language Models with GAN-Based LiDAR-to-RGB Synthesis for Autonomous Driving}

\author{Sven Kirchner\orcidlink{0009-0004-3845-6772}, Nils Purschke\orcidlink{0009-0008-9470-5795}, Ross Greer\orcidlink{0000-0001-8595-0379}, and Alois C. Knoll\orcidlink{0000-0003-4840-076X}, \IEEEmembership{Senior Member, IEEE}

\thanks{This work was supported by the Federal Ministry of Research, Technology and Space (BMFTR) as part of the CeCaS project, FKZ: 16ME0800K.}
\thanks{Sven Kirchner, Nils Purschke and Alois C. Knoll are with the Chair of Robotics, Artificial Intelligence and Real-time Systems,
        Technical University of Munich, 80333 Munich, Germany
        (email: sven.kirchner@tum.de)}%
\thanks{Ross Greer is with the Computer Science and Engineering Department,
        University of California Merced, Merced, CA 95343 USA}}

\maketitle


\begin{abstract}
Ensuring reliable autonomous operation when visual input is degraded remains a key challenge in intelligent vehicles and robotics. We present DepthVision, a multimodal framework that enables Vision--Language Models (VLMs) to exploit LiDAR data without any architectural changes or retraining. DepthVision synthesizes dense, RGB-like images from sparse LiDAR point clouds using a conditional GAN with an integrated refiner, and feeds these into off-the-shelf VLMs through their standard visual interface. A Luminance-Aware Modality Adaptation (LAMA) module fuses synthesized and real camera images by dynamically weighting each modality based on ambient lighting, compensating for degradation such as darkness or motion blur. This design turns LiDAR into a drop-in visual surrogate when RGB becomes unreliable, effectively extending the operational envelope of existing VLMs. We evaluate DepthVision on real and simulated datasets across multiple VLMs and safety-critical tasks, including vehicle-in-the-loop experiments. The results show substantial improvements in low-light scene understanding over RGB-only baselines while preserving full compatibility with frozen VLM architectures. These findings demonstrate that LiDAR-guided RGB synthesis is a practical pathway for integrating range sensing into modern vision–language systems for autonomous driving.
\end{abstract}




\section{Introduction}
\label{sec:Introduction}

Intelligent vehicles and autonomous driving systems rely on accurate environment perception, prediction, and planning to ensure safe decision making and control. Classic autonomy stacks process raw sensor data through modular pipelines—covering perception, motion prediction, and trajectory planning—to reconstruct the scene and forecast dynamic agent behavior \cite{sekkat2024amodalsynthdrive,  dal2025joint, greer2021trajectory}. These modular systems have demonstrated high robustness across diverse driving scenarios, yet overall performance remains constrained by upstream sensing quality and cross-module coordination \cite{paden2016survey, chen2022milestones}.

Recent developments in autonomous vehicle research have focused on enhancing each module within the traditional stack. For perception, multimodal fusion of LiDAR, radar, and cameras has improved reliability under challenging weather and illumination conditions \cite{xiang2023multi}. Prediction networks leverage deep spatio–temporal models to capture agent interaction dynamics, while planning components increasingly incorporate uncertainty and risk awareness for safe decision making in complex urban environments \cite{huang2022survey, mustafa2024racp}. Complementary efforts explore synthetic dataset generation and simulation frameworks to address data scarcity in rare or safety-critical cases \cite{song2023synthetic}. These advances have significantly strengthened the classical stack, but integration across modules remains limited. This motivates the exploration of unified, learning-based architectures that connect perception and control more directly \cite{teng2022hierarchical}.

Transformer-based architectures, leveraging the Self-Attention and Multi-Head Attention mechanisms introduced in \cite{vaswani2017}, have demonstrated the capacity to directly map raw vision inputs (e.g., camera images) and natural language instructions to robot actions \cite{brohan2022}. Unified sequence models with object-centric, multimodal prompting further expand the range of tasks that can be solved within a single model architecture, supporting general-purpose manipulation and interaction \cite{Jiang2023}. 
This has led to significant improvements in perception, prediction, and planning tasks through Vision–Language Models (VLM) and Vision-Language-Action models (VLA), including the introduction of end-to-end pipelines that map sensor input directly to control signals \cite{zhou2024vision, cui2024survey}.
Embodied multimodal language models (EMLMs) extend large pretrained language models (LLMs) with continuous sensor streams from robotic platforms, thereby enabling complex closed-loop control capabilities \cite{Driess2023}. 

These developments critically depend on sufficiently rich visual information to perceive relevant objects and scene dynamics and to inform downstream decision-making processes. Despite these advances, the rapid development of VLA models raises fundamental questions about the trade-offs, costs and limitations of current multimodal sensing strategies. While camera images provide dense semantic information, additional sensing modalities—such as LiDAR—enhance spatial awareness and physical obstacle detection, especially in low-light or poor-visibility conditions where cameras may struggle, though this comes at the cost of increased hardware complexity and expense \cite{feng2020deep}. Vision Transformers (ViTs) scale favorably with model and dataset size, given abundant training data \cite{Zhai2022}. However, the availability of web-scale multimodal datasets—spanning images and text—has predominantly benefited vision-based learning, as vast quantities of annotated visual data are readily available through online repositories, social media and public datasets \cite{Alayrac2022}. In contrast, LiDAR data remain scarce at scale, given the inherent costs and logistical barriers to mass collection and labeling of point clouds. Unlike images or videos, which are easily generated and shared by consumers, LiDAR scans and dense depth point clouds are rarely produced or published in large, diverse quantities. As a result, current foundation models overwhelmingly favor the visual modality, limiting the robustness of models deployed in real-world tasks that require accurate three-dimensional spatial reasoning. This challenge is increasingly relevant in robotics domains such as autonomous driving, where robust and reliable multimodal sensor fusion remains an open research question \cite{ChenBoyuan2024, gu2024clft}. 

Therefore, developing methods for processing multimodal sensors that balance the strengths and limitations of vision- and range-based data is a critical next step for the field. To this end, we propose DepthVision: a robust, multimodal framework that enables Vision–Language Models to maintain high performance, even when the real camera signal is degraded, unavailable or unreliable, by leveraging LiDAR sensing as an alternative modality.
By synthesizing RGB images from LiDAR data and adaptively fusing them with real camera input, DepthVision substantially improves scene understanding in challenging conditions such as low light, occlusion, or sensor failure. The major contributions of our research can be summarized as follows:
\begin{itemize}
    \item We introduce DepthVision, a multimodal framework that integrates LiDAR into Vision–Language models (VLMs) by synthesizing RGB images from sparse LiDAR data, enabling robust vision–language reasoning for autonomous driving when camera input is degraded or unavailable.
    \item We design a conditional GAN with a U\!-Net generator, PatchGAN discriminator, and lightweight residual refiner that produces realistic, geometry-consistent images under sparse depth by iteratively reducing artifacts and sharpening structure.
    \item We propose Luminance-Aware Modality Adaptation (LAMA) with two fusion strategies—global (full) and local (pixelwise)—that blend real and synthesized images based on scene luminance, down-weighting unreliable RGB in low light while preserving compatibility with frozen VLMs.
    \item We demonstrate on CARLA and nuScenes that zero-shot VLMs augmented with DepthVision achieve significant night-time gains on safety-critical perception and VQA-style tasks. Ablations and vehicle-in-the-loop tests confirm robustness and practical deployability under adverse illumination.
\end{itemize}

The rest of this paper is organized as follows: Section \ref{sec:related_work} reviews the most relevant literature, including multimodal sensing and fusion, vision–language models, and GAN-based image translation. Section \ref{sec:method} presents the proposed \emph{DepthVision} method, detailing LiDAR preprocessing, the GAN+refiner architecture, the LAMA fusion strategy, and VLM integration. Section \ref{sec:experiments} describes the experimental setup including the real vehicle integration and reports results on CARLA and nuScenes. Finally, Section \ref{sec:conclusion} concludes with key findings, limitations, and directions for future work.


\section{Related Work}
\label{sec:related_work}

In this section, we review key advances in multimodal sensor processing for autonomous driving, motivating LiDAR–camera integration as a means to enhance perception reliability. We also summarize recent progress in vision–language modeling and GAN-based image generation that informs our approach to robust multimodal scene understanding.

\subsection{Multimodal Perception for Autonomous Driving}\label{subsec:Multimodal Perception for Autonomous Driving}
Modern autonomy stacks must accurately perceive their surroundings—objects, layout, and motion—to enable reliable scene understanding and safe operation. Because road scenes vary widely in weather, illumination, and traffic, perception must remain robust to degradation and out-of-distribution conditions. A common strategy to achieve such robustness is sensor redundancy, i.e., fusing complementary modalities so that one sensor can compensate when another fails \cite{xiong2023lxl,zhang2022robust}. Fusion is typically organized at the data, feature, or decision level, with designs explicitly targeting robustness to single-sensor failures in perception and tracking \cite{wang2023multi}. Beyond this taxonomy, concrete LiDAR–camera fusion approaches have demonstrated task-level gains; for example, early-fusion detectors with a shared voxelized backbone improve both accuracy and efficiency in 3D detection by injecting RGB features into point-level representations \cite{wen2021fast}. LiDAR–camera fusion frameworks for road detection and scene layout enhance geometric reliability \cite{gu20183}, while joint calibration and synchronization tools improve cross-sensor alignment \cite{domhof2021joint}. Feature-level fusion learns modality-specific encoders (image and point/voxel/BEV) and aligns them via concatenation, cross-attention, or transformers; such designs strengthen geometric consistency for road layout and scene parsing, provided precise calibration and timing are maintained \cite{gu20183,domhof2021joint}. Decision-level fusion combines independent RGB and LiDAR predictors at the output stage to improve reliability under partial failures and distribution shifts \cite{wang2023multi}. At the same time, recent analyses note that even range sensors can degrade (e.g., on dark surfaces or low-reflectivity targets), motivating explicit sensor-quality assessment within the fusion loop \cite{shahbeigi2024novel}.

\subsection{Vision–Language Models in Autonomous Driving}\label{subsec:Vision–Language–Action Models in Robotics}
Recent advances in intelligent vehicles highlight the growing role of Vision-Language Models enabling richer semantic reasoning over fused inputs—supporting more informed perception and decision-making in complex scenes while bridging robotic perception, reasoning and control \cite{ma2024dolphins}. These models are being applied across core robotic tasks: zero-shot trajectory generation \cite{kwon2024}, multimodal task planning \cite{wake2024}, adaptive task replanning \cite{mei2024}, navigation \cite{song2024} and resource-efficient robotic manipulation \cite{wen2025}. Although targeting different robotic domains, all approaches depend critically on robust scene understanding. This has motivated research into multimodal sensor fusion and the enrichment of RGB images with depth data, with the aim of improving perception, even in challenging conditions. Collectively, these works demonstrate that reliable multimodal perception is fundamental to deploying VLM- and VLA-based systems in autonomous driving, particularly for safety-critical operation in unstructured real-world environments.

\subsection{Vision–Language Models and Multimodality}\label{subsec:Vision–Language Models and multimodality}
Large-scale Vision–Language Models have demonstrated strong performance on tasks that require joint reasoning over images and text. These models extend pretrained language backbones by integrating visual encoders, enabling complex tasks such as multimodal dialogue, image captioning, visual question answering and instruction following. A foundational enabler of modern VLMs is the Vision Transformer (ViT)\cite{dosovitskiy2020}, which adapts the self-attention mechanism originally developed for natural language processing to the visual domain. Since a naive application of self-attention at the pixel level would be computationally prohibitive, ViTs restructure an image $\mathbf{x} \in \mathbb{R}^{H \times W \times C}$ into a sequence of flattened 2D patches $\mathbf{x}_p \in \mathbb{R}^{N \times (P^2 C)}$, where $(H, W)$ is the image resolution, $C$ the number of channels, \(P \times P\) the patch size and $N = HW / P^2$ the resulting sequence length. This allows the image to be processed in the same way as a token sequence in text, unlocking scalable multimodal attention. While conventional computer vision pipelines focus on predicting a fixed set of object classes, VLMs instead learn to align visual and linguistic modalities in a shared embedding space \cite{radford21}. For example, models like VisualBERT \cite{li2019} jointly process visual and textual inputs by embedding region-based visual features and text tokens, which are then fed into a single transformer encoder. The model is pretrained using masked language modeling and an image–text matching objective, enabling it to learn rich cross-modal representations. Some vision–language models, such as FLAVA \cite{Singh2022}, extend this idea by employing a fully shared transformer encoder across both modalities, supporting unified learning over image, text and multimodal inputs. Recent work has begun adapting VLMs to incorporate 3D data. Chen et al. \cite{Chen2024} leverage monocular metric depth estimation to lift large-scale 2D internet images into metric 3D point clouds, enhancing the 3D scene understanding and question answering capabilities of vision–language models. Yang et al. \cite{Yang2025} propose integrating LiDAR data with large language models by flattening the 3D point cloud along the z-axis to generate a bird’s-eye-view (BEV) feature map, which is then used as a prompt input to a view-aware transformer module within the LLM.

\begin{figure*}[htb]
    \vspace{1em}
    \centering
    \includegraphics[width=0.96\textwidth]{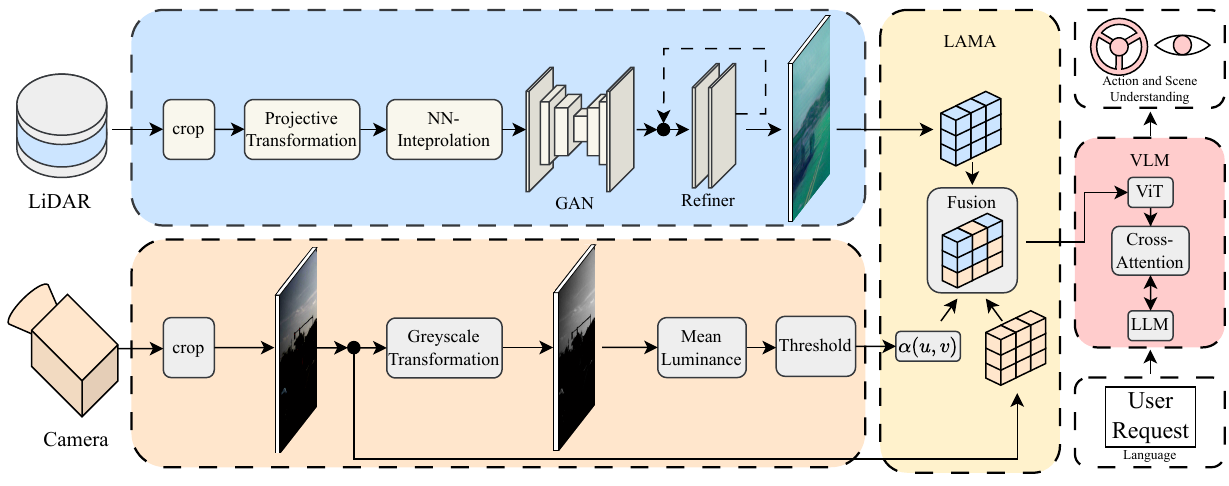}  %
    \caption{DepthVision architecture with four components: LiDAR processing (blue), RGB preprocessing (orange), luminance-aware fusion via LAMA (yellow), and the Vision–Language Model (red). LiDAR inputs are converted into RGB-like images through a GAN–refiner pipeline, and the VLM performs action or scene understanding. The system adaptively selects or blends modalities based on scene brightness to provide a robust visual input to the VLM.}
  \label{fig:DepthVision_Architecture}
\end{figure*}

\subsection{Generative Adversarial Networks and Refiner}\label{subsec:Generative Adversarial Networks and refiner}
Generative Adversarial Networks (GANs) were first introduced by Goodfellow et al.\cite{Goodfellow2014} as a framework for generating realistic synthetic data through an adversarial game between a generator and a discriminator. The Deep Convolutional GAN (DCGAN)\cite{radford2016} extended this idea by using deep convolutional layers to improve stability and visual quality in image generation tasks. For image-to-image translation, architectures such as the U-Net \cite{Ronneberger2015} are widely used because their encoder–decoder structure with skip connections helps preserve spatial details. The pix2pix framework \cite{Isola2017} applied conditional GANs to learn mappings between paired image domains, making it suitable for tasks like translating depth maps into RGB images. To improve the quality of generated images, Shrivastava et al. \cite{Shrivastava2017} proposed adding a refiner network that cleans and enhances synthetic outputs, making them more realistic. This is especially important for ill-posed problems where the input data is sparse or incomplete. For example, generating a three-channel RGB image from a single-channel depth map requires the network to hallucinate missing color and texture information. One effective strategy for handling such under-constrained mappings is to use iterative refinement methods \cite{Adler2017}. Instead of generating the final image in a single step, the network produces an initial estimate and then improves it in multiple stages using a learned or fixed update rule. This step-by-step approach helps the model converge to a higher-quality result with better structural consistency and fewer artifacts.


\section{Method}
\label{sec:method}

\subsection{Method Overview} \label{subsec: Method Overview}

We propose a novel method for multimodal scene understanding, outlined in Fig.~\ref{fig:DepthVision_Architecture}. DepthVision augments frozen vision--language models with a LiDAR-guided visual stream that remains reliable under poor illumination. Given a calibrated RGB--LiDAR pair, we first project LiDAR points into the image plane to obtain a sparse depth map aligned to the camera. A conditional GAN with a U\!-Net generator and a PatchGAN discriminator translates this sparse map into a dense, RGB-like view; a lightweight residual refiner iteratively suppresses artifacts and sharpens geometry. We then perform luminance-aware fusion between the real RGB frame and the synthesized view: a global or pixelwise gate, computed from RGB intensities, down-weights unreliable RGB in dark regions without modifying the downstream VLM. The fused image is tokenized by a ViT-style patch embedder and concatenated with text tokens for multimodal reasoning.

\subsection{LiDAR Preprocessing} \label{sec: LiDAR preprocessing}

To align the LiDAR with the RGB camera we use a standard pinhole model with
known extrinsics and intrinsics. We assume zero skew, separate focal lengths
\(f_x,f_y\), and that lens distortion has been removed from the image; i.e.,
projection is performed in the undistorted camera frame.

\begin{figure}[!h]
    \centering
    \includegraphics[width=\linewidth]{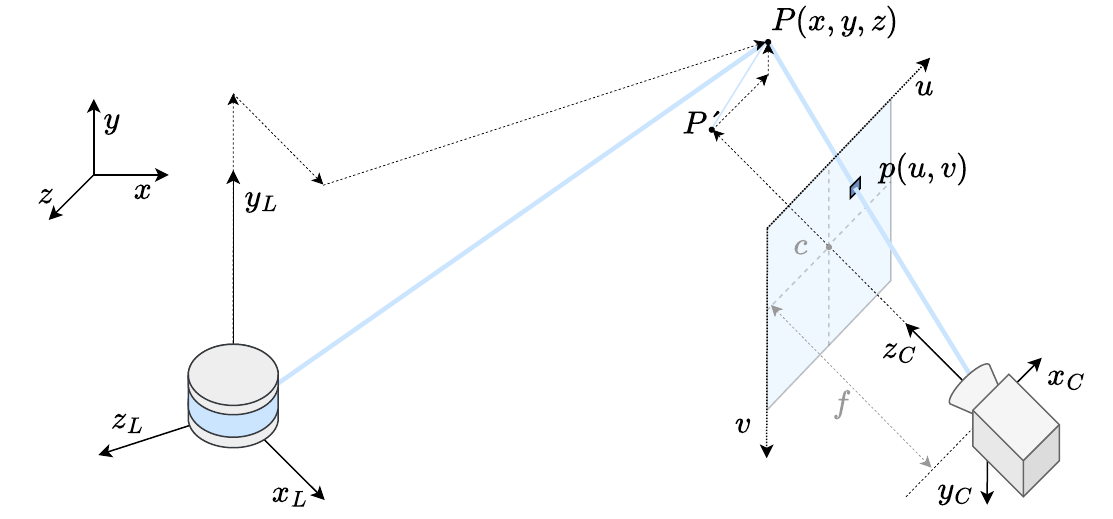}
    \caption{LiDAR-to-camera projection geometry. A 3D point measured in the
    LiDAR frame \((x_L,y_L,z_L)\) is transformed into the camera frame
    \((x_C,y_C,z_C)\) using the extrinsics \(\mathbf{T}_{C\leftarrow L}\) and
    then projected to pixel coordinates \(p(u,v)\) by the intrinsics
    \(\mathbf{K}\).}
    \label{fig:camera_LiDAR}
\end{figure}

Let a LiDAR point in homogeneous coordinates be
\(\tilde{\mathbf{P}}_L=[x_L,y_L,z_L,1]^\mathsf{T}\).
It is first transformed into the camera coordinate frame by the extrinsics
\(\mathbf{T}_{C\leftarrow L}\in\mathbb{R}^{4\times 4}\):
\begin{equation}
\tilde{\mathbf{P}}_C =
\mathbf{T}_{C\leftarrow L}\,\tilde{\mathbf{P}}_L,\qquad
\mathbf{T}_{C\leftarrow L}=
\begin{bmatrix}
\mathbf{R} & \mathbf{t} \\
\mathbf{0}^\mathsf{T} & 1
\end{bmatrix},
\end{equation}
where \(\mathbf{R}\in\mathbb{R}^{3\times 3}\) and
\(\mathbf{t}\in\mathbb{R}^{3}\).
Let \(\mathbf{P}_C=[X_C,Y_C,Z_C]^\mathsf{T}\) denote the inhomogeneous camera
coordinates. The perspective projection to the image plane is
\begin{equation}
s\,\tilde{\mathbf{p}} =
\mathbf{K}\,
\begin{bmatrix}
\mathbf{R} & \mathbf{t}
\end{bmatrix}
\tilde{\mathbf{P}}_L
=
\mathbf{K}\,\mathbf{P}_C,\qquad
\tilde{\mathbf{p}}=[u,v,1]^\mathsf{T},
\label{eq:projection}
\end{equation}
with projective depth \(s=Z_C\) and intrinsic matrix
\begin{equation}
\mathbf{K}=
\begin{bmatrix}
f_x & 0   & c_x \\
0   & f_y & c_y \\
0   & 0   & 1
\end{bmatrix}.
\end{equation}
Expanding \eqref{eq:projection} yields the pixel coordinates
\begin{equation}
u = f_x \frac{X_C}{Z_C} + c_x,\qquad
v = f_y \frac{Y_C}{Z_C} + c_y.
\label{eq:uvcoords}
\end{equation}

Only points that are visible and inside the image are retained:
\begin{equation}
Z_C>0,\qquad 0\le u < W,\qquad 0\le v < H.
\end{equation}
If multiple 3D points project to the same pixel, we keep the one with the
smallest depth (z-buffering): \(Z_C=\min Z_C\) at that pixel. The resulting depth map is cropped to $600\times600$ pixels centered at the principal point $(c_x,c_y)$.

The valid projections form a sparse depth map \(D\) on the image lattice.
To match the downstream processing resolution of $512\times512$ and obtain a denser 2D representation while preserving depth discontinuities, we resample this cropped map onto a $512\times512$ grid and apply nearest-neighbor interpolation over the set of valid pixels 
$\Omega \subset \{0,\ldots,W\!-\!1\}\!\times\!\{0,\ldots,H\!-\!1\}$.
We use nearest-neighbor interpolation because it preserves sharp depth discontinuities and does not hallucinate intermediate depths, preventing “bleeding” across object boundaries that bilinear/bicubic schemes introduce under sparsity. It is also computationally cheap and robust when $\Omega$ is very sparse. Although NN can yield blocky artifacts, these are later corrected by the GAN+refiner; edge-aware alternatives (e.g., bilateral/guided interpolation) are slower and can be biased by unreliable RGB in low light—the regime we target.
\begin{equation}
\hat{D}(u,v) =
D\!\left(
\underset{(i,j)\in\Omega}{\arg\min}
\,\!\sqrt{(u-i)^2+(v-j)^2}
\right).
\label{eq:interpolation}
\end{equation}

\begin{figure*}[htb]
    \vspace{1em}
    \centering
    \includegraphics[width=0.96\textwidth]{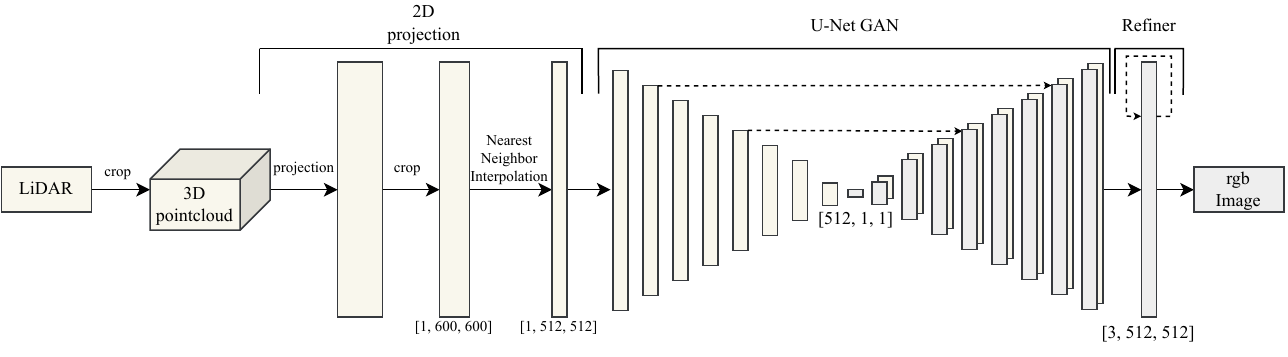}  %
  \caption{LiDAR-to-RGB synthesis pipeline in DepthVision. The 3D point cloud is projected and interpolated into a 2D depth map, which is translated to an RGB image by a U-Net GAN and iteratively refined by a lightweight refiner network.}
  \label{fig:LiDAR_processing_pipeline}
\end{figure*}

\subsection{GAN and Refiner Setup} \label{sec: GAN setup}

To synthesize RGB images from projected LiDAR inputs, we use a conditional Generative Adversarial Network (GAN) (See Fig.~\ref{fig:LiDAR_processing_pipeline}) based on the \textit{pix2pix} framework \cite{Isola2017}. All RGB images are normalized to the range $[-1, 1]$ across channels. The generator follows a U\!-Net encoder--decoder architecture with skip connections between corresponding layers to preserve spatial detail and edge structure. This design enables the model to learn a structured mapping from sparse LiDAR projections to dense RGB domains while maintaining geometric consistency.

The generator takes a single-channel LiDAR projection of size $512 \times 512$ as input and outputs a three-channel RGB image of the same resolution. This synthetic image serves as input to the subsequent VLM, providing robustness in multimodal tasks even when real RGB data is degraded or unavailable (e.g., due to poor lighting).

The discriminator adopts the PatchGAN formulation, assessing image realism over local $70 \times 70$ patches rather than the full image at once. This local supervision encourages high-frequency fidelity and fine texture generation.

To further enhance visual quality,  we introduce a lightweight refiner module to perform residual correction through iterative refinement. The refiner is defined as a compact, fully convolutional, network comprising three convolutional layers with ReLU activations and a final $\tanh$ output layer to bound the residual output within $\left[-1,\,1\right]$. It consists of a simple three-layer $3\times3$ CNN with 64 feature channels in the hidden layers and shared weights across all refinement steps. At each iteration, the current RGB estimate and LiDAR projection are fused and passed through the refiner, which predicts a residual term subsequently added to the RGB input. Three refinement iterations are applied, progressively reducing artifacts and improving structural coherence.

The output of the generation process is first fused with the RGB image using weighted fusion. This ensures that the vision language model downstream benefits from a dense, RGB-like representation, even in scenarios where real camera data is unreliable or missing.

\subsection{Luminance-Aware Modality Adaptation (LAMA)} \label{sec: Luminance-Aware modality adaption}

To achieve robust scene understanding under varying illumination conditions, we introduce two luminance-aware modality adaptation strategies. These methods are designed to maintain reliability when the RGB modality becomes unreliable, such as in low-light environments. The first, full fusion strategy, computes the mean luminance of the entire RGB image and adjusts the fusion weights globally based on this value. In contrast, the pixelwise fusion strategy computes luminance at each individual pixel, enabling spatially adaptive blending that emphasizes the GAN-generated image in darker regions while preserving real RGB content in well-lit areas. Given an RGB input image $\mathbf{I}_{\text{RGB}} \in \mathbb{R}^{H \times W \times 3}$, pixel values are normalized to the range $[0, 1]$ and converted to grayscale using the Rec. 709 luminance formula:
\begin{equation}
\mathbf{I}_{\text{gray}}(u, v) = 0.2126 \cdot R(u, v) + 0.7152 \cdot G(u, v) + 0.0722 \cdot B(u, v)
\end{equation}

For the full-fusion strategy, the mean luminance of the grayscale image determines the scene brightness:

\begin{equation}
L_{\text{mean}} = \frac{1}{HW} \sum_{u=1}^{W} \sum_{v=1}^{H} \mathbf{I}_{\text{gray}}(u, v)
\end{equation}

Two threshold values, $L_{\text{low}}$ and $L_{\text{high}}$, define a luminance band that guides modality fusion:

\begin{itemize}
    \item If $L_{\text{mean}} \leq L_{\text{low}}$, the RGB input is considered too dark and the GAN image is used.
    \item If $L_{\text{mean}} \geq L_{\text{high}}$, the RGB image is considered reliable and used directly.
    \item If $L_{\text{mean}} \in (L_{\text{low}}, L_{\text{high}})$, we compute a weighted linear blend between the two:
\end{itemize}

\begin{equation}
\alpha = \frac{L_{\text{mean}} - L_{\text{low}}}{L_{\text{high}} - L_{\text{low}}}
\end{equation}
\begin{equation}
\mathbf{I}_{\text{fused}} = \alpha \cdot \mathbf{I}_{\text{RGB}} + (1 - \alpha) \cdot \mathbf{I}_{\text{GAN}}
\end{equation}

Alternatively, pixelwise blending computes $\alpha(u,v)$ locally from the luminance at each pixel:

\begin{equation}
\alpha(u, v) = \text{clamp}\left(\frac{\mathbf{I}_{\text{gray}}(u, v) - L_{\text{low}}}{L_{\text{high}} - L_{\text{low}}}, 0, 1\right)
\end{equation}

where $\text{clamp}(x, 0, 1)$ constrains the value of $x$ to the range $[0, 1]$, ensuring valid blending weights. This allows spatially adaptive fusion, in which dark regions rely more on the GAN image while bright areas retain real RGB information.

\begin{figure}[]
    \centering
    \includegraphics[width=\linewidth]{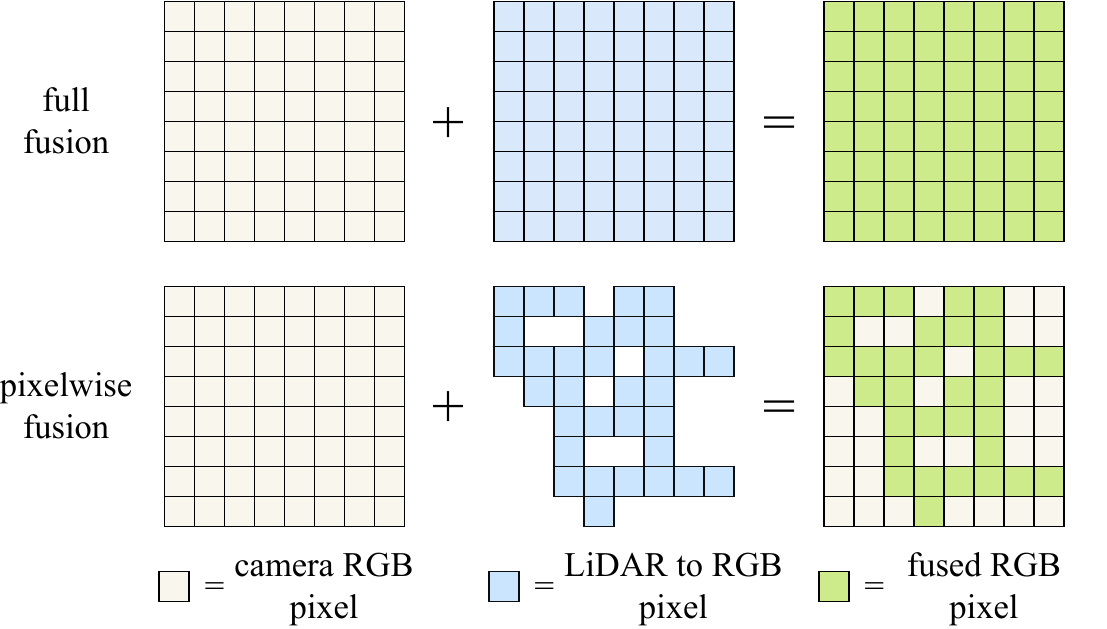}
    \caption{
    Overview of the two luminance-aware fusion strategies used for combining RGB and LiDAR modalities. \textbf{(a) full fusion} applies a single global weighting factor to both inputs, determined by the mean scene luminance of the RGB input. \textbf{(b) pixelwise fusion} computes a spatially varying weight map based on per-pixel luminance, allowing the model to blend information locally and adaptively. This enables fine-grained emphasis on LiDAR in poorly lit regions while still leveraging RGB information in well-exposed areas.
    }
    \label{fig:fusion}
\end{figure}

In practice, we set the luminance thresholds to $L_{\text{low}} = 0.15$ and $L_{\text{high}} = 0.35$, enabling smooth and continuous transition between modalities under varying illumination conditions. Fig.~\ref{fig:fusion} illustrates the two main fusion strategies: full-image (global) fusion and pixelwise (local) blending. By leveraging luminance cues, the system adaptively reduces reliance on unreliable RGB input in challenging scenarios—such as nighttime scenes, tunnels, or occluded environments—thereby enabling robust, context-aware multimodal perception.

\subsection{VLM Integration} \label{sec: general architecture}

The final stage of the DepthVision pipeline fuses multimodal sensor data into a unified VLM, enabling high-level scene understanding across diverse environmental conditions.

\paragraph*{Inputs} The system takes the following inputs:

\begin{itemize}
    \item \textbf{LiDAR point cloud:} A set of 3D points $\mathbf{P} = \{\mathbf{p}_i\} \subset \mathbb{R}^3$, projected into a single-channel tensor $\mathbf{I}_{\text{LiDAR}} \in \mathbb{R}^{1 \times H \times W}$.
    \item \textbf{RGB image:} A three-channel tensor $\mathbf{I}_{\text{RGB}} \in \mathbb{R}^{3 \times H \times W}$ captured by the camera, resized to $H = W = 512$ and normalized to $[0,1]$.
    \item \textbf{Text input:} A sequence of token embeddings $\mathbf{T} = \{\mathbf{t}_1, \dots, \mathbf{t}_n\} \subset \mathbb{R}^{d}$, where $n$ denotes the number of language tokens and $d$ the embedding dimension.
\end{itemize}

\paragraph*{Modality Adaptation} Based on LAMA, as described in Section \ref{sec: Luminance-Aware modality adaption}, we dynamically adapt the visual input to the VLM depending on the reliability of the RGB sensor. When the RGB image is well-exposed, it is passed directly to the model. Under poor illumination, we substitute the RGB view with a synthetic LiDAR-to-RGB image generated by the GAN. For intermediate lighting conditions, the two modalities are linearly blended using a luminance-dependent weighting factor. The resulting fused image $\mathbf{I}_{\text{fused}}$ is converted to standard RGB format and can be directly processed by frozen VLMs without additional fine-tuning.

\paragraph*{Transformer Input} The fused image $\mathbf{I}_{\text{fused}}$ is tokenized through a Vision Transformer (ViT) patch embedding module to produce a sequence of $N = \frac{H \cdot W}{P^2}$ image tokens, where $P$ is the patch size. These visual tokens are concatenated with the language token embeddings to form a joint input sequence:

\begin{equation}
\mathbf{Z} = \mathrm{Transformer}\left( \left[ \mathrm{PatchEmbed}(\mathbf{I}_{\text{fused}}) \, ; \, \mathbf{T} \right] \right)
\end{equation}

This luminance-driven adaptation occurs entirely outside the VLM, maintaining full compatibility with frozen model weights while enhancing perceptual robustness.

\paragraph*{Robustness Through Redundancy.} By allowing the system to switch its emphasis between RGB and LiDAR based on observed luminance, DepthVision achieves robustness in challenging visual conditions such as:

\begin{itemize}
    \item low illumination (e.g., nighttime or tunnels)
    \item visual artifacts (e.g., glare, shadows, overexposure)
    \item partial sensor failure or dropout
\end{itemize}

\paragraph*{Output.} The VLM outputs task-specific predictions such as scene descriptions, object categories, and navigation decisions. This design makes DepthVision adaptable for a wide range of downstream applications in autonomous driving and mobile robotics.


\section{Experiments}
\label{sec:experiments}

\subsection{Experimental Setup}
\begin{figure}[!h]
    \centering
    \includegraphics[width=\linewidth]{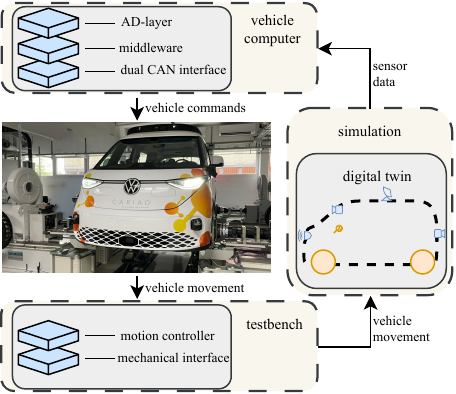}
    \caption{Vehicle-in-the-loop setup for integration testing. The vehicle computer receives synthetic sensor data from a simulation hosting a digital twin in a virtual environment and returns vehicle commands via the AD (autonomous driving) layer, middleware, and a redundant CAN interface. These commands actuate the vehicle, which operates on a physical testbench through a mechanical interface and motion controller. The measured vehicle movement is fed back to the simulation, thus closing the loop.}
    \label{fig:vehicle_integration}
\end{figure}

We evaluate DepthVision both in the standalone quality of LiDAR$\to$RGB synthesis and the end-to-end impact on multimodal scene understanding with VLM integration. Experiments span simulation and real-world data and include a vehicle-in-the-loop setup to assess integration feasibility (Fig.~\ref{fig:vehicle_integration}).

The vehicle is controlled via a computer with the hardware setup shown in Table \ref{tab:hardware}.

\begin{table}[H]
\centering
\caption{Hardware setup of the vehicle computer.}
\label{tab:hardware}
\renewcommand{\arraystretch}{1.1}
\begin{tabular}{ll}
\hline
\textbf{Name} & Vehicle Computer \\ \hline
\textbf{CPU}  & AMD TRP 5955WX \\ 
 & 16 Cores at 2.7 GHz \\ \hline
\textbf{GPU} & RTX 4090 24 GB VRAM \\ \hline
\textbf{RAM} & 64 GB DDR4-3200 RAM \\ \hline
\end{tabular}
\end{table}

\subsection{Design and Implementation}
To ensure reproducibility, we provide detailed descriptions of the model architectures and training configurations used in our experiments. The LiDAR-to-RGB synthesis pipeline is based on the conditional GAN framework introduced in pix2pix \cite{Isola2017}. We have extended this framework by adding an extra convolutional layer and integrating a lightweight residual refiner module which enhances the quality and structural consistency of the generated images. All input images are resized to $512 \times 512$ pixels and normalized to the range $[-1, 1]$. Training is conducted using mixed-precision arithmetic on NVIDIA GPUs to improve performance and memory efficiency. The GAN and refiner network are trained jointly.

For multimodal scene understanding, we evaluate state-of-the-art Vision–Language Models VLMs without fine-tuning. Specifically, we include \textit{LLaVA-1.6-Mistral-7B} \cite{li2024} and \textit{Qwen2-VL-7B-Instruct} \cite{Wang2024}, both of which offer strong performance on open-ended visual reasoning and instruction-following tasks. Additionally, we wrap the Vision–Language Models in an architecture inspired by the \textit{EMMA} architecture \cite{hwang2024}, implemented using the official \textit{OpenEMMA} codebase \cite{Xing2024}.

\subsection{Datasets and Training Details}

\paragraph*{Sensor Setup.} 
We train two separate DepthVision networks: one using simulation-based data from CARLA~\cite{Dosovitskiy17} and another using real-world data from the nuScenes dataset~\cite{Caesar2020}. Table~\ref{tab:sensor_specs} outlines the key sensor specifications for both configurations. 

\begin{table}[H]
\centering
\caption{Sensor specifications for simulation and real-world datasets.}
\label{tab:sensor_specs}
\renewcommand{\arraystretch}{1.1}
\begin{tabular}{@{}p{3cm}p{2.5cm}p{2.5cm}@{}}
\toprule
\textbf{Parameter}   & \textbf{CARLA}       & \textbf{nuScenes}   \\ \midrule
Camera - Position   & x: 1.8\,m, z: 1.7\,m\textsuperscript{*} & Roof front          \\
Camera - Resolution        & 800$\times$600 px  & 1600$\times$900 px  \\
Camera - Field of View  & 90° horiz.          & $\sim$70° horiz.    \\
Camera - Frame Rate        & 5\,Hz                 & 12\,Hz               \\
LiDAR - Position    & x: 1.8\,m, z: 1.7\,m\textsuperscript{*} & Roof center         \\
LiDAR - Channels    & 64                    & 32                  \\
LiDAR - FOV         & 90° horiz., 74° vert. & 360° horiz., 40° vert. \\
LiDAR - Range             & 100\,m                & 70–120\,m            \\
LiDAR - Frame Rate  & 10\,Hz                & 20\,Hz               \\
LiDAR - Points / Second & 256k                  & $\sim$300k  
\\
Training - Samples & 5744 & 28130
\\ \bottomrule
\addlinespace[0.5ex]
\multicolumn{3}{l}{\small\textsuperscript{*}Positions are relative to the vehicle center.} \\
\end{tabular}
\end{table}

To ensure consistent lighting conditions during training, we filter out samples with a mean luminance below 0.4. This ensures that the LiDAR-to-RGB synthesis network learns to generate primarily day-like images, thereby avoiding overfitting to poorly illuminated or low-information scenes. While the CARLA environment features a simplified test map with reduced complexity, the nuScenes-based model is trained and evaluated on more diverse and challenging real-world scenarios. The front camera and LiDAR are calibrated using known extrinsic and intrinsic parameters. For nuScenes, we densify the sparse point cloud by fusing up to three sequential LiDAR sweeps using vehicle odometry before projection.

\paragraph*{Training and Hyperparameters}

All components of the LiDAR-to-RGB synthesis pipeline—the generator, discriminator and refiner—are trained jointly using a combination of adversarial and reconstruction objectives. The total loss consists of a binary cross-entropy adversarial term and an $L_1$ reconstruction loss scaled by a factor of 100, encouraging both structural accuracy and color fidelity.

Training is performed using the Adam optimizer~\cite{Diederik2015} with $\beta_1 = 0.5$, $\beta_2 = 0.999$ and a learning rate of $2 \times 10^{-4}$. A batch size of 16 is used and mixed-precision arithmetic is employed to improve computational efficiency and memory usage. The training process is parallelized across two NVIDIA RTX 4090 GPUs. The full set of training hyperparameters is summarized in Table~\ref{tab:training_hyperparams}.

\begin{table}[H]
\centering
\caption{Training hyperparameters.}
\label{tab:training_hyperparams}
\renewcommand{\arraystretch}{1.1}
\begin{tabular}{ll}
\toprule
\textbf{Parameter}               & \textbf{Value} \\ \midrule
Optimizer                        & Adam ($\beta_1 = 0.5$, $\beta_2 = 0.999$) \\
Learning rate                    & $2 \times 10^{-4}$ \\
Batch size                       & 16 \\
Epochs                           & 200 \\
Loss functions                   & Adversarial + $L_1$ \\
$L_1$ loss weight ($\lambda$)    & 100 \\
Input scaling                    & RGB pixel values normalized to [$-1$, $1$] \\
Luminance threshold              & 0.4 (low-light filter) \\
\bottomrule
\end{tabular}
\end{table}

\subsection{Simulation Data}
\begin{figure}[!h]
    \centering
    \includegraphics[width=\linewidth]{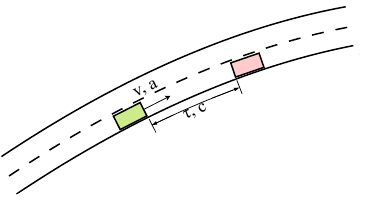}
    \caption{The adaptive cruise control (ACC) as defined in ISO 15622 (Intelligent Transport Systems — Adaptive cruise control systems — Performance requirements and test procedures). The time gap is denoted by $\tau$, the clearance by $c$, the ego vehicle speed by $v$, and the acceleration by $a$.}
    \label{fig:test_scenario}
\end{figure}

In order to evaluate the LiDAR-to-RGB synthesis as well as the VLM performance in low-light settings, we designed a dedicated night-time evaluation environment in CARLA. Scenes were generated using fixed geometry, calibrated sensors and minimal ambient illumination to mimic real-world visibility degradation, while ensuring precise alignment between the LiDAR and RGB modalities. Notably, the DepthVision model was trained exclusively on daytime data, enabling it to generate daytime-like images in previously unseen night-time conditions.
\paragraph*{Vehicle integration}
To evaluate the impact of LiDAR-guided vision on safety-critical perception and decision making, we integrate the complete DepthVision perception module into the AD-layer of our software stack (see Fig. \ref{fig:vehicle_integration}) and deploy it on the vehicle computer (see Table~\ref{tab:hardware}). We then design an adaptive cruise control (ACC) scenario (Fig.~\ref{fig:test_scenario}), following ISO~15622 \cite{ISO15622}, and execute it in the CARLA simulation environment under night-time conditions.

In this experiment, we isolate the LiDAR-to-RGB component of DepthVision. The synthesized images are fed into a lightweight YOLOP detector \cite{wu2022yolop} to localize the lead vehicle, and a simple PID controller maintains the desired following distance. Braking is triggered solely from the detection signal and controller output. We run the scenario once with night-time camera images and once with LiDAR-to-RGB synthesized images. Analyzing the system behavior allows us to evaluate the latency, stability, and robustness of our LiDAR-guided pipeline under low-light conditions. In the test scenario, the ego vehicle is required to follow a lead vehicle traveling at a constant speed of $10\,\mathrm{m/s}$ while maintaining a clearance of $c = 10\,\mathrm{m}$. Importantly, the AD-software interacts with the real vehicle electronics, our middleware, and the control interface, demonstrating that the proposed approach is fully compatible with real-world automotive hardware.

\begin{figure}[]
  \centering
  \includegraphics[width=\linewidth]{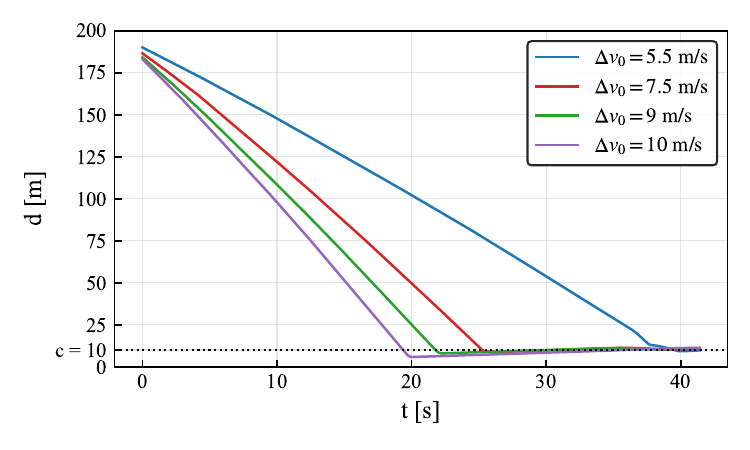}
  \caption{Ego-to-lead vehicle distance $d(t)$ in the ISO~15622 ACC scenario for different initial relative velocities $\Delta v_0$. The dashed line denotes the required clearance $c = 10\,\mathrm{m}$.}
  \label{fig:distance_figure}
\end{figure}

\begin{figure}[]
  \centering
  \includegraphics[width=\linewidth]{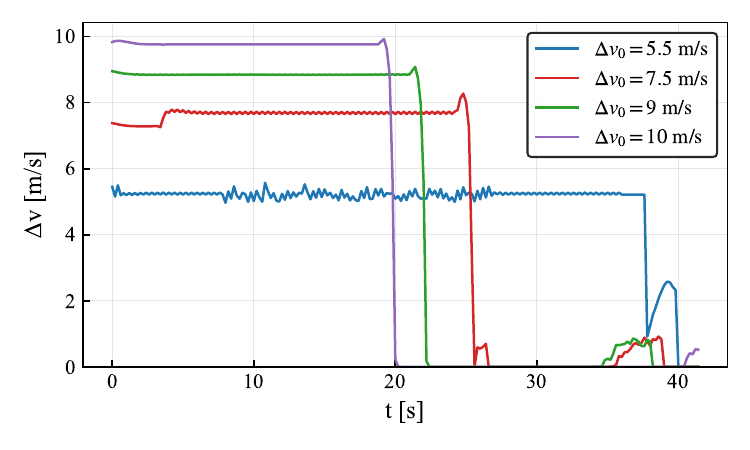}
  \caption{Relative velocity $\Delta$v(t) between ego and lead vehicle for the same ISO~15622 ACC scenarios. DepthVision enables safe braking to $\Delta$v $\approx$0 before the clearance c is violated, even at higher $\Delta v_0$.}
  \label{fig:speed_figure}
\end{figure}

We evaluate DepthVision for different initial relative velocities
$\Delta v_0 \in \{5.5,\,7.5,\,9,\,10\}\,\mathrm{m/s}$, representing increasingly critical closing speeds. With LiDAR-to-RGB synthesis, the ego vehicle remains stable throughout the scenario, decelerates smoothly as $\Delta v$ approaches zero (Fig.~\ref{fig:speed_figure}), and converges to a steady following distance close to the required $c = 10\,\mathrm{m}$, as shown in Fig.~\ref{fig:distance_figure}. In contrast, when using camera-only input, the YOLOP-based perception stack fails to reliably detect the environment at night, causing the controller to leave the lane before even approaching the lead vehicle, and preventing any meaningful distance regulation.

\begin{figure*}[!t]
    \centering
    \includegraphics[width=\textwidth]{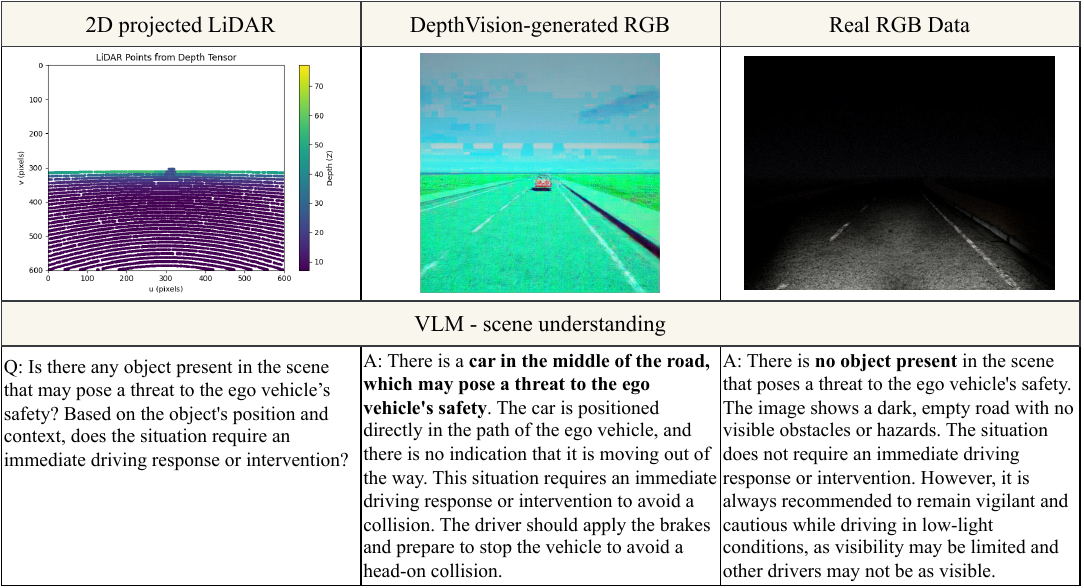}
    \caption{Simulation evaluation. A nighttime highway scenario shows a vehicle ahead at a safety-critical distance requiring braking. Due to inactive lights, the vehicle is not visible in the real image. In contrast, the DepthVision-generated image clearly reveals the vehicle, enabling the VLM (QWEN) to reason correctly and suggest braking.}
    \label{fig:qualitative_simulation}
\end{figure*}

\paragraph*{Vision-Question Answering (VQA)}
Having demonstrated that DepthVision enables stable closed-loop control in a real vehicle setup, we next examine its primary advantage: the ability to make Vision–Language Models LiDAR-aware. To this end, we evaluate how well VLMs can perform visual question answering (VQA) when provided with LiDAR-guided synthesized imagery. LiDAR scans of the night-time environment were processed through the DepthVision pipeline to generate synthetic RGB images. As Fig.~\ref{fig:qualitative_simulation} shows, despite the sparsity of the input, the model successfully preserves critical semantic and structural cues, such as vehicle contours and scene layout. The generator produces plausible textures and colours based on depth geometry, while the refiner improves edge sharpness and reduces artifacts. Compared to raw night-time RGB captures, the synthetic images improve visibility, compensating for sensor noise and glare using information learned from well-lit scenes. By converting sparse LiDAR data into a dense RGB-like representation, DepthVision allows the Vision-Language Model to process LiDAR information via its standard visual interface. This enables the VLM to interpret the scene, even when the real camera signal is unreliable.

\subsection{Baseline for Real-World VQA} 

To enable a controlled comparison with our proposed DepthVision framework, we construct a baseline system focused on front-view multimodal scene understanding using large-scale VLMs. This baseline is modeled after NuScenes-QA \cite{Qian2024}, but is adapted to single-frame, real-world settings with calibrated LiDAR and RGB inputs.

\paragraph*{Dataset Composition} We curated a subset from the nuScenes validation set \cite{Caesar2020}, selecting day and night scenes from the front camera perspective. Each sample includes a single RGB image, the corresponding LiDAR sweep and complete calibration metadata. The LiDAR sweep was cropped to fit the camera's field of view. The evaluation set consists of 400 VQA tasks.

\paragraph*{Question Types and Annotation} Each image is paired with manually authored question–answer (QA) pairs, that can be categorized into three task types, plus an overall accuracy metric:
\begin{itemize}
    \item \textbf{Existence (Safety Critical):} Questions that ask for the number of safety-critical objects in the scene (e.g., vehicles in close proximity that may pose a collision risk), requiring accurate object detection and counting.
    \item \textbf{Count:} Questions that require counting the number of objects in a specific class (e.g., cars, trucks, pedestrians or motorcycles).
    \item \textbf{Object Classification:} Questions that ask for the classification of an object, often based on its spatial relation to other objects or scene context.
    \item \textbf{Acc:} The overall Top-1 accuracy, computed as the average across all three tasks above.
\end{itemize}
Ground truth labels are derived from a combination of human annotations and nuScenes 3D bounding boxes, cross-validated via projected sensor views for spatial consistency. All models are evaluated in zero-shot mode without fine-tuning, using GAN-generated images and corresponding natural language queries. Evaluation metrics include Top-1 accuracy across QA categories. This setup provides a reproducible benchmark for understanding scenes across multiple modalities and facilitates the quantitative evaluation of the impact of DepthVision.

\subsection{Real-World Performance}

\begin{table*}[t]
\centering
\caption{Top-1 accuracy of vision–language models using camera and LiDAR-to-RGB synthesis under varying fusion strategies and lighting conditions.}
\label{tab:method_comparison}
\renewcommand{\arraystretch}{1.2}
\begin{tabular}{@{}ll*{15}{c}@{}}
\toprule
\multirow{2}{*}{\textbf{VLM}} & \multirow{2}{*}{\textbf{DataSource (fusion strategy)}} 
& \multicolumn{3}{c}{\textbf{Exist (Safety Critical)}} 
& \multicolumn{3}{c}{\textbf{Count}} 
& \multicolumn{3}{c}{\textbf{Object}} 
& \multicolumn{3}{c}{\textbf{Acc}} \\
\cmidrule(lr){3-5} \cmidrule(lr){6-8} \cmidrule(lr){9-11} \cmidrule(lr){12-14}
& & Day & Night & All & Day & Night & All & Day & Night & All & Day & Night & All \\ 
\midrule
QWEN & Camera-only & 54.8 & 53.3 & 54.1 & 86.9 & 88.3 & 87.6 & 94.7 & 74.7 & 84.7 & 78.8 & 72.1 & 75.5 \\ 
QWEN & Camera + LiDAR (full) & 54.8\textsuperscript{†} & \textbf{66.7} & \textbf{60.8} & 86.9\textsuperscript{†} & 84.5 & 85.7 & 94.7\textsuperscript{†} & 73.1 & 83.9 & 78.8\textsuperscript{†} & 74.8 & \textbf{76.8} \\ 
QWEN & Camera + LiDAR (pixelwise) & 47.6 & 44.4 & 46.0 & 77.4 & 85.5 & 81.5 & 63.2 & \textbf{80.7} & 72.0 & 62.7 & 70.2 & 66.5 \\ 
\specialrule{0.05pt}{0pt}{0pt}
LLAVA & Camera-only & 83.0 & 57.8 & 70.4 & 79.9 & 78.3 & 79.1 & 68.4 & 57.7 & 63.1 & 77.1 & 64.6 & 70.9 \\ 
LLAVA & Camera + LiDAR (full) & 83.0\textsuperscript{†} & \textbf{73.3} & \textbf{78.2} & 79.9\textsuperscript{†} & \textbf{83.9} & 81.9 & 68.4\textsuperscript{†} & 53.8 & 61.1 & 77.1\textsuperscript{†} & 70.3 & \textbf{73.7} \\ 
LLAVA & Camera + LiDAR (pixelwise) & 82.9 & 60.0 & 71.5 & 85.4 & 79.4 & 82.4 & 55.0 & \textbf{61.6} & 58.3 & 74.4 & 67.0 & 70.7 \\ 
\bottomrule
\addlinespace[0.5ex]
\multicolumn{14}{l}{\small\textsuperscript{†}Same as Camera-only. Fusion was not applied during daytime scenes when using full fusion.} \\
\end{tabular}
\end{table*}


We evaluate the effectiveness of our LiDAR-augmented visual reasoning framework on real world data by comparing baseline Vision–Language Model (VLM) performance under three visual input conditions: (i) camera-only (RGB) and two versions of DepthVision (ii) RGB + LiDAR via full fusion and (iii) RGB + LiDAR via pixelwise fusion. Table \ref{tab:method_comparison} shows the quantitative results for three reasoning tasks — existence (safety-critical), object counting and object classification — under both day and night conditions.

\paragraph*{LiDAR Augmentation Improves Low-Light Scene Understanding} As shown in Table \ref{tab:method_comparison}, augmenting RGB inputs with LiDAR-generated imagery significantly enhances VLM performance under low-luminance conditions. In particular, fusing the refiner GAN-generated LiDAR image via full fusion substantially improves the accuracy of safety-critical object detection at night for Qwen2-VL-7B-Instruct (from 53.3\% to 66.7\%, a \textbf{13.4\%} absolute improvement) and for LLaVA-1.6-Mistral-7B, an improvement of \textbf{15.5\%}. This pattern is consistent across models, indicating that the additional structural information from LiDAR mitigates the degradation in RGB image quality at night.

\paragraph*{Fusion Strategy Matters: Full vs. Pixelwise Fusion.} The experimental results indicate that, in general, full fusion outperforms pixel-wise fusion across most tasks. Notably, pixel-wise fusion yields a relative improvement of up to \textbf{6 \%} in object classification, suggesting its effectiveness in preserving fine-grained object details under low-light conditions by locally enhancing dark regions. However, this localized blending can introduce inconsistencies in well-lit scenes, occasionally degrading performance during daytime scenarios. In contrast, full fusion provides a more stable and globally consistent representation across varying lighting conditions

\paragraph*{Luminance-Aware Performance Trends}

\begin{figure}[]
  \centering
  \includegraphics[width=\linewidth]{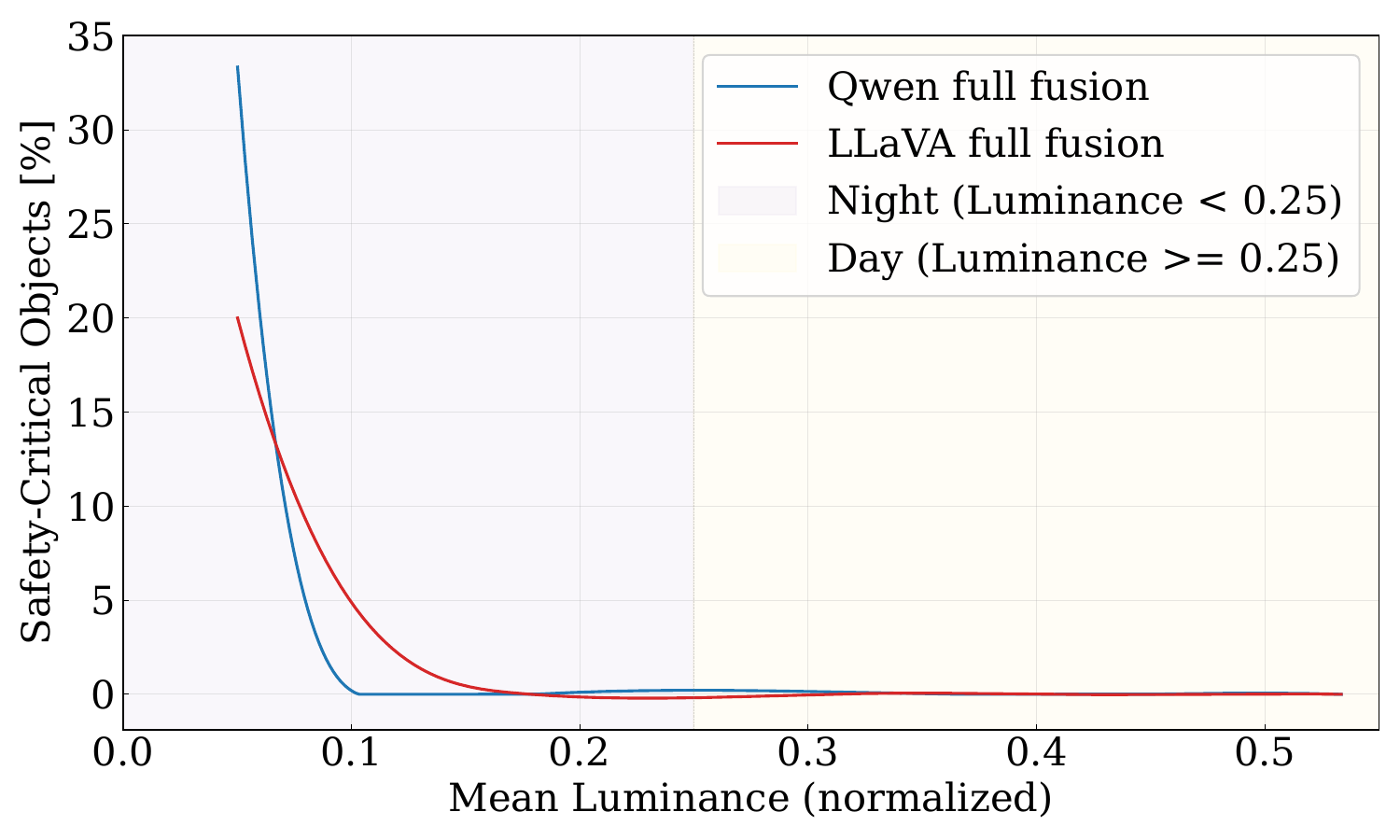}
  \caption{Relative accuracy improvement in detecting safety-critical objects when using DepthVision with full weighted fusion compared to camera-only input.}
  \label{fig:improvement_safety_critical}
\end{figure}

\begin{figure}[]
  \centering
  \includegraphics[width=\linewidth]{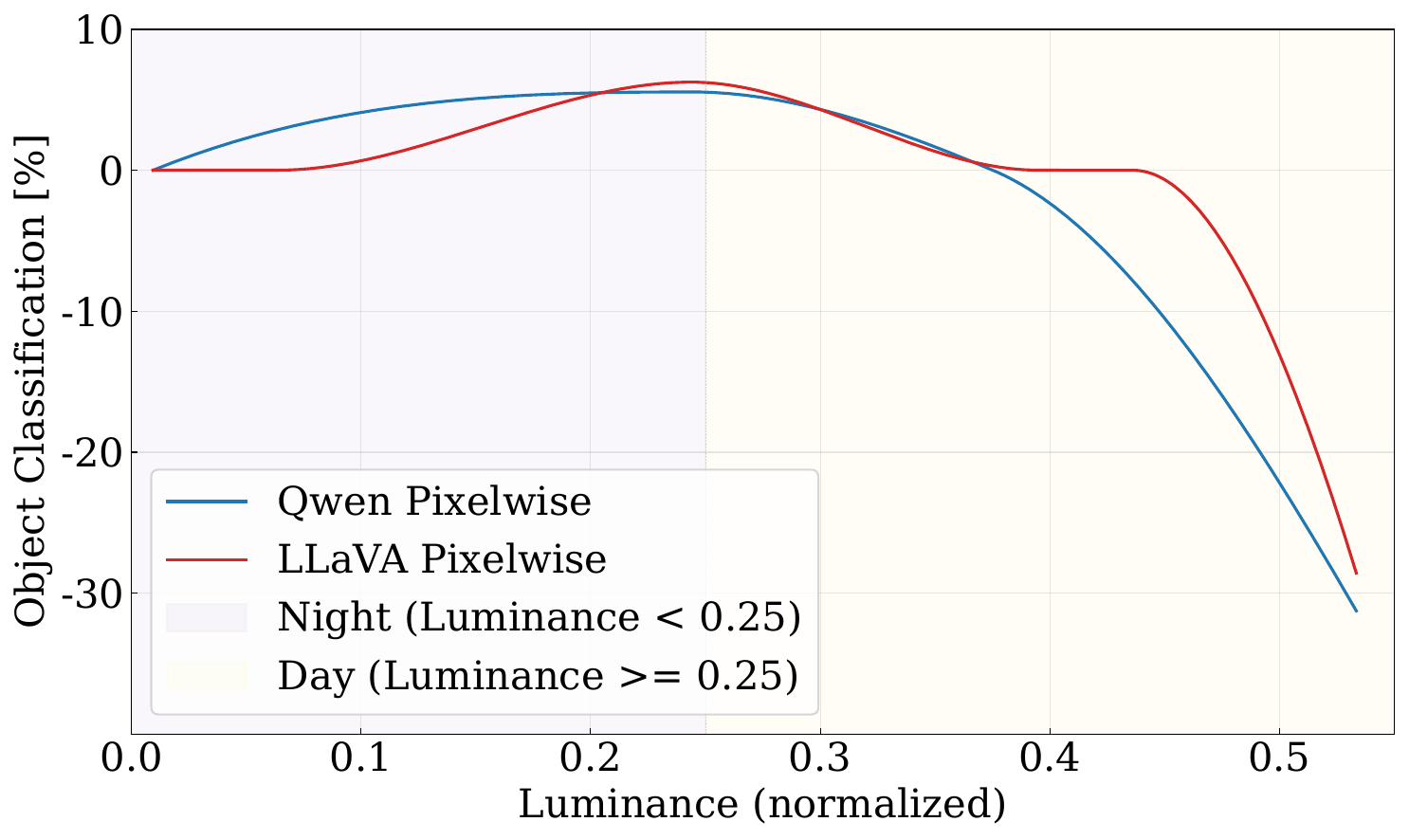}
  \caption{Relative improvement in object classification accuracy with pixelwise weighted fusion compared to camera-only.}
  \label{fig:improvement_object_classification}
\end{figure}

To better understand the effect of ambient lighting on model performance, we compute the mean luminance of each input image and analyze its correlation with VLM accuracy. Fig.~\ref{fig:improvement_safety_critical} and Fig.\ref{fig:improvement_object_classification} show separate performance trends for each task (Safety Critical and Object Classification), comparing RGB-only inputs with fused RGB+LiDAR inputs. The results reveal several key patterns:

\begin{itemize}
\item A consistent improvement in performance under low-light (nighttime) conditions when using fused inputs.
\item A performance drop during the daytime when using pixel-wise fusion, which is likely due to suboptimal blending in uniformly well-lit scenes. This may be because frozen VLMs are not trained to process such fused inputs, which leads to distribution shifts. Disabling fusion above a defined luminance threshold can mitigate this issue.
\item Increasing performance differences between RGB-only and fused inputs as luminance decreases, highlighting the complementary role of LiDAR in low visibility scenarios.
\end{itemize}

\paragraph*{Task-Specific Gains} The most significant impact of LiDAR augmentation is seen in safety-critical detection tasks, where structural cues, such as object boundaries and contours, are vital for identifying obscured or dimly lit vehicles and pedestrians. The Count and Object Classification tasks also demonstrate greater robustness, particularly in challenging night-time conditions.
These results reinforce the central hypothesis of this work: that real-time LiDAR-guided RGB synthesis can enhance downstream multimodal reasoning in low-visibility scenarios, without requiring retraining or architectural modifications to existing Vision–Language Models.


\section{Conclusion}
\label{sec:conclusion}
We presented DepthVision, a robust multimodal framework that enables vision-language models to exploit LiDAR data effectively, thereby facilitating reliable vision-language reasoning in challenging sensing conditions. At its core, the system uses a GAN-based LiDAR-to-RGB synthesis network with an integrated refiner to generate realistic images from sparse depth data. These synthetic views are then combined with real camera input via Luminance-Aware Modality Adaptation (LAMA), which fuses the two types of input adaptively based on scene brightness. This approach enhances perception and end-to-end pipelines in scenarios where RGB input is compromised, such as at night or in low-luminance environments, while remaining fully compatible with pre-trained vision-language models. Experiments on simulated and real-world data, together with vehicle integration and closed-loop testing, confirm that our method significantly improves performance in safety-critical and object-level tasks, particularly at night.


\bibliographystyle{IEEEtran}
\bibliography{literature}

\begin{IEEEbiography}
[{\includegraphics[width=1in,height=1.25in,clip,keepaspectratio]{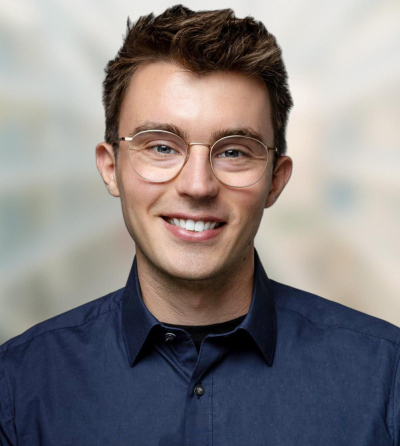}}]{Sven Kirchner}
received the B.Sc. and M.Sc. degrees in mechanical engineering from RWTH Aachen
University, Germany. In 2023, he joined
the Chair of Robotics, Artificial Intelligence, and
Real-time Systems at the Technical University of Munich, Munich, Germany, as a Ph.D. student. \\
His research interests include probabilistic robotics and data-driven approaches employing machine learning and deep learning for safety-critical systems. His work focuses on the design of robust, end-to-end multimodal models to enable safe and reliable autonomous driving.
\end{IEEEbiography}

\begin{IEEEbiography}[{\includegraphics[width=1in,height=1.25in,clip,keepaspectratio]{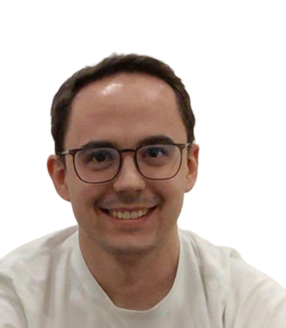}}]{Nils Purschke}
received his B.Sc.\ and M.Sc.\ degrees in computer science from the Technical University of Darmstadt, Germany, in 2021 and 2023. In 2023 he started his Ph.D. with the Technical University of Munich and is currently working toward his degree in the area of computer science.\\
He is a Research Associate with the Chair of Robotics, Artificial Intelligence and Real-time Systems under the lead of Prof. Knoll. His research areas include automated software-engineering, artificial intelligence, and automotive systems.\\
Parallel to his scientific work, he has been working in the industry as a Software Engineer in IT security since 2020.
\end{IEEEbiography}

\begin{IEEEbiography}[{\includegraphics[width=1in,height=1.25in,clip,keepaspectratio]{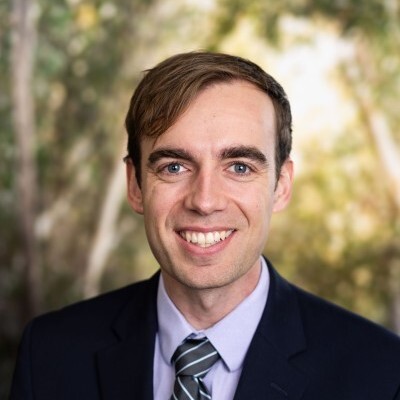}}]{Ross Greer}
is an Assistant Professor in the Department of Computer Science \& Engineering at the University of California, Merced. His research explores questions in computer vision and machine learning related to safe, human-interactive autonomous systems, especially focused on handling cases which sit on the long-tail distribution of novel and risky real-world scenarios. His research emphasizes active and representation learning, vision-language and multimodal information, human-robot interaction, and autonomous perception, prediction, and planning.
\end{IEEEbiography}

\begin{IEEEbiography}[{\includegraphics[width=1in,height=1.25in,clip,keepaspectratio]{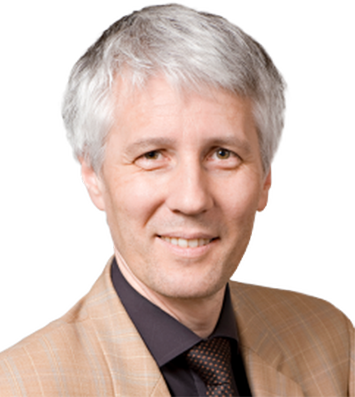}}]{Alois C. Knoll}
(Senior Member) received the diploma (M.Sc.) degree in electrical/communications
engineering from the University of Stuttgart, Stuttgart, Germany, in 1985, and the Ph.D. (summa cum laude) degree in computer science from the Technical University of Berlin (TU Berlin), Berlin, Germany, in 1988. He served on the Faculty of the Computer Science Department, TU Berlin till 1993. He joined the University of  Bielefeld, Bielefeld, Germany, as a Full Professor and was the Director of the Technical Informatics Research Group till 2001. Since 2001, he has been a Professor with the Department of Informatics, Technical University of Munich (TUM), Munich, Germany. He was also on the Board of Directors of the Central Institute of Medical Technology, TUM (IMETUM). From 2004 to 2006, he was the Executive Director of the Institute of Computer Science, TUM. His research interests include cognitive, medical, and sensor-based robotics, multiagent systems, data fusion, adaptive systems, multimedia information retrieval, model-driven development of embedded systems with applications to automotive software and electric transportation, and simulation systems for robotics and traffic. Between 2007 and 2009, he was a
member of the EU’s highest advisory board on information technology and the
Information Society Technology Advisory Group, and a member of its subgroup
on Future and Emerging Technologies. In this capacity, he was actively involved
in developing the concept of the EU’s FET Flagship projects.
\end{IEEEbiography}

\end{document}